\documentclass[runningheads]{llncs}
\usepackage[T1]{fontenc}
\usepackage{microtype}
\usepackage{wrapfig}
\usepackage{amsfonts}
\usepackage{algpseudocode}  % For pseudocode commands
\usepackage{algorithmicx}  % For additional algorithmic commands
\usepackage{amsmath}  % For math symbols
\usepackage[ruled,vlined]{algorithm2e}

\usepackage{amsmath}
\usepackage{graphicx}
\begin{document}

\title{Wasserstein Adaptive Value Estimation \\ for Actor-Critic Reinforcement Learning}
\titlerunning{Wasserstein Adaptive Value Estimation}
\author{Ali Baheri\inst{1}\orcidID{0000-0002-5613-0192} \and
Zahra Shahrooei\inst{1}\orcidID{0009-0008-7899-0514} \and
Chirayu Salgarkar\inst{1}\orcidID{0009-0007-7291-1312}}
\authorrunning{A. Baheri et al.}
\institute{Rochester Institute of Technology, Rochester NY 14623, USA 
\email{\{akbeme,zs9580,cms8111\}@rit.edu}}
\maketitle              % typeset the header of the contribution
\begin{abstract}
We present {\textbf{{W}}}asserstein \textbf{{A}}daptive \textbf{{V}}alue \textbf{{E}}stimation for Actor-Critic (WAVE), an approach to enhance stability in deep reinforcement learning through adaptive Wasserstein regularization. Our method addresses the inherent instability of actor-critic algorithms by incorporating an adaptively weighted Wasserstein regularization term into the critic's loss function. We prove that WAVE achieves the $\mathcal{O}\left(\frac{1}{k}\right)$ convergence rate for the critic's mean squared error and provide theoretical guarantees for stability through Wasserstein-based regularization. Using the Sinkhorn approximation for computational efficiency, our approach automatically adjusts the regularization based on the agent's performance. Theoretical analysis and experimental results demonstrate that WAVE achieves superior performance compared to standard actor-critic methods.

\end{abstract}
\section{Introduction}
Reinforcement learning (RL) has become a powerful framework for solving complex decision-making tasks across domains such as robotics, finance, healthcare, and autonomous systems \cite{kober2013reinforcement,hambly2023recent,razzaghi2024survey,yu2021reinforcement}. Among the various RL algorithms, actor-critic methods have gained significant attention due to their capacity to handle large state and action spaces, particularly when combined with deep neural networks \cite{grondman2012survey}. Actor-critic algorithms are advantageous because they use both value-based and policy-based learning: the critic estimates a value function to assess the policy, while the actor uses this assessment to optimize the policy itself. Despite their theoretical strengths, actor-critic algorithms often suffer from instability and slow convergence during training, particularly in continuous and high-dimensional environments. These issues are primarily due to the variance in value function estimates, correlated updates between the actor and critic, and approximation errors inherent in using neural networks. To mitigate these challenges, regularization techniques have been proposed to improve the stability and robustness of RL algorithms. Existing regularization approaches, such as weight decay or entropy regularization, provide some degree of stability but do not adequately address the temporal and distributional consistency in value estimates—two critical factors in achieving stable learning in actor-critic frameworks \cite{cayci2022finite}. In this context, optimal transport theory and the Wasserstein metric present a compelling avenue for improving the stability of value function estimates by quantifying discrepancies between distributions. 

This paper proposes a simple approach that integrates adaptive Wasserstein regularization into the critic’s loss function of actor-critic algorithms to address the challenges of instability and slow convergence. Specifically, we propose a regularization term based on the Sinkhorn approximation of the Wasserstein distance between consecutive value function estimates. By penalizing large deviations between successive Q-value distributions, this regularization term promotes temporal consistency, effectively reducing variance in the critic’s updates and leading to smoother learning dynamics. The contributions of this paper are threefold:

\begin{itemize}
    \item We develop a framework that incorporates an adaptive Wasserstein regularization term into the critic's loss function of general actor-critic algorithms.

    \item We provide theoretical analysis demonstrating that the adaptive Wasserstein regularization enhances the convergence properties of actor-critic algorithms. 

    \item Through experiments on standard RL benchmarks with continuous action spaces, we validate the effectiveness of the proposed method.
\end{itemize}
\subsection{Related Work}
The integration of Wasserstein distance into RL frameworks is a significant development in recent years \cite{richemond2017wasserstein,abdullah2018reinforcement,metelli2019propagating,shahrooei2024optimal,baheri2023risk,baheri2024synergy}. The Wasserstein distance, known for providing a geometry-aware topology superior to existing f-divergences, has found particular utility in both unsupervised and robust RL paradigms \cite{hou2020robust}. For instance, the Wasserstein robust RL (WR2L) framework formalizes robust RL as a max-min game with Wasserstein constraints, particularly effective in both low and high-dimensional control tasks \cite{abdullah2019wasserstein}. This approach has demonstrated significant improvements in handling environmental uncertainties and dynamics variations. The incorporation of Wasserstein distance into actor-critic architectures has led to several developments. The Wasserstein actor-critic (WAC) approach employs approximate Q-posteriors to represent epistemic uncertainty and uses Wasserstein barycenters for uncertainty propagation across the state-action space \cite{likmeta2023wasserstein}. This integration has proven particularly effective in addressing value estimation errors and improving sample efficiency. Recent theoretical work has established important connections between Wasserstein distance and RL optimization. Research has shown that the Wasserstein metric can effectively measure the distance between different policies' induced distributions, providing a more robust foundation for policy optimization. Mean-field analysis has further demonstrated that neural actor-critic methods with Wasserstein constraints can achieve global optimality at sublinear rates \cite{zhang2021wasserstein}. In the context of unsupervised RL, the Wasserstein distance has been employed to maximize the diversity of learned policies. The Wasserstein unsupervised RL (WURL) framework directly maximizes the distance between state distributions induced by different policies, offering advantages over mutual information-based methods \cite{he2022wasserstein}. This approach enables agents to explore state spaces more effectively and develop diverse skills that can be applied to downstream tasks through hierarchical learning.
\section{Methodology}
Actor-critic algorithms are a class of RL methods that employ two function approximators: an actor, which selects actions based on a policy, and a critic, which evaluates the policy by estimating value functions. This framework combines the advantages of both policy-based and value-based methods. We consider a Markov decision process (MDP) defined by the tuple $(\mathcal{S}, \mathcal{A}, P, r, \gamma)$ where $\mathcal{S}$ is the state space, $\mathcal{A}$ is the action space (which can be continuous or discrete), $P\left(s^{\prime} \mid s, a\right)$ is the state transition probability, $r: \mathcal{S} \times \mathcal{A} \rightarrow \mathbb{R}$ is the reward function, and $\gamma \in[0,1)$ is the discount factor. The objective is to find an optimal policy $\pi^*$ that maximizes the expected discounted return
$
J\left(\theta^\pi\right)=\mathbb{E}_\pi\left[\sum_{t=0}^{\infty} \gamma^t r\left(s_t, a_t\right)\right],
$
where $a_t \sim \pi\left(a_t \mid s_t ; \theta^\pi\right)$ and $s_{t+1} \sim P\left(s_{t+1} \mid s_t, a_t\right)$.
\subsection{Actor and Critic Updates}
In the actor update, the policy parameters $\theta^\pi$ are updated in the direction that maximizes the expected return. Using the policy gradient theorem

$$
\nabla_{\theta^\pi} J\left(\theta^\pi\right)=\mathbb{E}_{s \sim d^*, a \sim \pi}\left[\nabla_{\theta^\pi} \log \pi\left(a \mid s ; \theta^\pi\right) Q^\pi(s, a)\right],
$$
The critic approximates $Q^\pi\left(s, a ; \theta^Q\right)$, parameterized by $\theta^Q$, and is updated by minimizing the temporal difference (TD) error

$$
L_{\mathrm{TD}}\left(\theta^Q\right)=\mathbb{E}_{\left(s, a, r, s^{\prime}\right)}\left[\left(Q^\pi\left(s, a ; \theta^Q\right)-y\right)^2\right]
$$
where $y=r+\gamma \mathbb{E}_{a^{\prime} \sim \pi}\left[Q^\pi\left(s^{\prime}, a^{\prime} ; \theta^Q\right)\right]$. To enhance the stability of the critic's learning process, we propose an adaptive regularization term based on the Wasserstein distance between consecutive value function estimates.
\begin{definition}[Wasserstein-$p$ Distance]
Given two probability measures $\mu$ and $\nu$ on a metric space $(\mathcal{X}, d)$, the Wasserstein-p distance is defined as:

\[
W_p(\mu, \nu)=\left(\inf _{\gamma \in \Gamma(\mu, \nu)} \int_{\mathcal{X} \times \mathcal{X}} d(x, y)^p d \gamma(x, y)\right)^{1 / p}
\]
where $\Gamma(\mu, \nu)$ is the set of all couplings of $\mu$ and $\nu$.

\end{definition}
Computing the exact Wasserstein distance can be computationally expensive. We employ the Sinkhorn distance, an entropic regularization of optimal transport, which provides an efficient approximation.

\begin{definition}[Sinkhorn Distance]
For empirical distributions $\hat{\mu}=\sum_{i=1}^n \delta_{x_i} / n$ and $\hat{\nu}=\sum_{j=1}^n \delta_{y_j} / n$, the Sinkhorn distance is

\[
W_{\varepsilon}(\hat{\mu}, \hat{\nu})=\min _{\pi \in \Pi(\hat{\mu}, \hat{\nu})} \sum_{i, j} \pi_{i j} c\left(x_i, y_j\right)-\varepsilon H(\pi),
\]
where $c\left(x_i, y_j\right)=\left\|x_i-y_j\right\|^2$ is the cost function, $H(\pi)=-\sum_{i, j} \pi_{i j} \log \pi_{i j}$ is the entropy of the coupling $\pi$, and $\varepsilon>0$ is the entropic regularization parameter.
\end{definition}

\begin{lemma}[Sinkhorn distance approximates Wasserstein distance]
The Sinkhorn distance $W_{\varepsilon}(\hat{\mu}, \hat{\nu})$ approximates the Wasserstein distance $W_1(\hat{\mu}, \hat{\nu})$ and is differentiable with respect to the input distributions.
\end{lemma}
We augment the critic's loss function with a Wasserstein regularization term

\[
L_{\mathrm{reg}}\left(\theta^Q\right)=L_{\mathrm{TD}}\left(\theta^Q\right)+\lambda_k W_{\varepsilon}\left(\hat{Q}_k, \hat{Q}_{k-1}\right)
\]
where $\lambda_k \geq 0$ is the adaptive regularization parameter at iteration $k$, and $\hat{Q}_k$ and $\hat{Q}_{k-1}$ are empirical distributions of the critic's $Q$-value estimates at iterations $k$ and $k-1$, respectively. The regularization parameter $\lambda_k$ is adapted based on the agent's performance to balance the bias-variance trade-off. Let $R_k$ denote the cumulative reward at episode $k$. We define the moving average of the cumulative reward over the last $M$ episodes:
$
\bar{R}_k=\frac{1}{M} \sum_{i=k-M+1}^k R_i .
$
The adaptive regularization parameter $\lambda_k$ is updated as

\[
\lambda_k= \begin{cases}\lambda_{\max }, & \text { if } \bar{R}_k \leq R_{\mathrm{th}} \\ \lambda_{\min }+\left(\lambda_{\max }-\lambda_{\min }\right) e^{-\alpha\left(\bar{R}_k-R_{\mathrm{th}}\right),} & \text { if } \bar{R}_k>R_{\mathrm{th}}\end{cases}
\]
where $\lambda_{\max }$ and $\lambda_{\min }$ are maximum and minimum regularization values, $R_{\mathrm{th}}$ is the performance threshold, and $\alpha>0$ controls the decay rate.
\begin{lemma}[$\lambda_k$ decrease]
   The adaptive scheme ensures that $\lambda_k$ decreases as the agent's performance improves, allowing the critic to focus more on minimizing the TD error. 
\end{lemma}
The critic parameters $\theta^Q$ are updated by minimizing the regularized loss function

\[
\theta_{k+1}^Q=\theta_k^Q-\eta_k \nabla_{\theta^Q} L_{\mathrm{reg}}\left(\theta_k^Q\right)
\]
where $\eta_k$ is the learning rate. The actor parameters $\theta^\pi$ are updated using the policy gradient

\[\nabla_{\theta^*} J\left(\theta^\pi\right)=\mathbb{E}_{s \sim d^*, a \sim \pi}\left[\nabla_{\theta^*} \log \pi\left(a \mid s ; \theta^\pi\right) Q^\pi\left(s, a ; \theta^Q\right)\right].\]

Algorithm~\ref{alg:wave-rl} presents the WAVE procedure, which augments actor-critic learning with adaptive Wasserstein regularization. The algorithm begins by initializing the actor and critic networks with parameters \(\theta^{\pi}\) and \(\theta^{Q}\) respectively, along with the initial regularization parameter \(\lambda_{\text{max}}\). For each episode, the agent interacts with the environment to collect trajectories consisting of state-action-reward transitions \((s_t, a_t, r_t, s_{t+1})\). The critic's parameters are then updated by minimizing the regularized loss \(L_{\text{reg}}(\theta^{Q})\), which incorporates both the standard temporal difference error and the Wasserstein distance between consecutive Q-value distributions. The actor parameters are subsequently updated using the policy gradient, exploiting the critic's value estimates. After each episode, the algorithm computes the cumulative reward \(R_k\) and updates the running average \(\bar{R}_k\), which is used to adaptively adjust the regularization parameter \(\lambda_k\) according to the agent's performance. 

\section{Main Theoretical Results}

\begin{theorem}[Convergence rate of the WAVE]
Under certain assumptions, the mean squared error (MSE) of the critic's Q-value estimates with adaptive Wasserstein regularization decreases at a rate of $\mathcal{O}\left(\frac{1}{k}\right)$, where $k$ is the iteration index. Consequently, the critic's parameter estimates $\theta_k^Q$ converge to the optimal parameters $\theta^{Q^*}$ at a rate of $\mathcal{O}\left(\frac{1}{\sqrt{k}}\right)$.    
\end{theorem}
\begin{proof}
We aim to establish that the critic's parameter error, given by \( \mathbb{E}\left[\left\|\theta_k^Q - \theta^{Q^*}\right\|^2\right] \), decreases at a rate of \( \mathcal{O}\left(\frac{1}{k}\right) \) under specific conditions. We begin by introducing some assumptions to facilitate the analysis. Assume that rewards are bounded such that \( |r(s, a)| \leq R_{\max} \) for all \( (s, a) \in \mathcal{S} \times \mathcal{A} \). The critic function \( Q(s, a; \theta^Q) \) is assumed to be Lipschitz continuous with respect to \( \theta^Q \), so that 
\[
\left\| Q(s, a; \theta_1^Q) - Q(s, a; \theta_2^Q) \right\| \leq L_Q \left\| \theta_1^Q - \theta_2^Q \right\|
\]
for any \( \theta_1^Q, \theta_2^Q \). The sequence of learning rates \( \{\alpha_k\} \) is defined as \( \alpha_k = \frac{a}{k} \) with \( a > 0 \), ensuring that \( \sum_{k=1}^\infty \alpha_k = \infty \) and \( \sum_{k=1}^\infty \alpha_k^2 < \infty \). Additionally, the gradients of the regularized loss function are bounded
\[
\left\| \nabla_{\theta^Q} L_{\mathrm{reg}}(\theta^Q) \right\| \leq G \quad \text{for all } \theta^Q.
\]
We also assume that the regularized loss function \( L_{\mathrm{reg}}(\theta^Q) \) is strongly convex with parameter \( m > 0 \), so that
\[
L_{\mathrm{reg}}(\theta_2^Q) \geq L_{\mathrm{reg}}(\theta_1^Q) + \nabla_{\theta^Q} L_{\mathrm{reg}}(\theta_1^Q)^T (\theta_2^Q - \theta_1^Q) + \frac{m}{2} \left\| \theta_2^Q - \theta_1^Q \right\|^2
\]
for any \( \theta_1^Q, \theta_2^Q \). The critic's parameters are updated at each iteration \( k \) as follows:
\[
\theta_{k+1}^Q = \theta_k^Q - \alpha_k \nabla_{\theta^Q} L_{\mathrm{reg}}(\theta_k^Q),
\]
where \( \nabla_{\theta^Q} L_{\mathrm{reg}}(\theta_k^Q) \) represents the gradient of the regularized loss function. We define the error vector as \( e_k = \theta_k^Q - \theta^{Q^*} \), where \( \theta^{Q^*} \) is the minimizer of \( L_{\mathrm{reg}}(\theta^Q) \). The error recursion can be written as:
\[
e_{k+1} = e_k - \alpha_k \nabla_{\theta^Q} L_{\mathrm{reg}}(\theta_k^Q).
\]
We analyze the expected squared norm of the error
\[
\mathbb{E}\left[\left\| e_{k+1} \right\|^2\right] = \mathbb{E}\left[\left\| e_k - \alpha_k \nabla_{\theta^Q} L_{\mathrm{reg}}(\theta_k^Q) \right\|^2\right].
\]
Expanding this expression yields:
\[
\mathbb{E}\left[\left\| e_{k+1} \right\|^2\right] = \mathbb{E}\left[\left\| e_k \right\|^2\right] - 2\alpha_k \mathbb{E}\left[ e_k^T \nabla_{\theta^Q} L_{\mathrm{reg}}(\theta_k^Q) \right] + \alpha_k^2 \mathbb{E}\left[ \left\| \nabla_{\theta^Q} L_{\mathrm{reg}}(\theta_k^Q) \right\|^2 \right].
\]
From the strong convexity of \( L_{\mathrm{reg}}(\theta^Q) \), we have
$
L_{\mathrm{reg}}(\theta_k^Q) - L_{\mathrm{reg}}(\theta^{Q^*}) \geq \frac{m}{2} \left\| e_k \right\|^2.
$
Since \( \nabla_{\theta^Q} L_{\mathrm{reg}}(\theta^{Q^*}) = 0 \), it follows that
\[
\nabla_{\theta^Q} L_{\mathrm{reg}}(\theta_k^Q) = \nabla_{\theta^Q} L_{\mathrm{reg}}(\theta_k^Q) - \nabla_{\theta^Q} L_{\mathrm{reg}}(\theta^{Q^*}).
\]
Applying the Lipschitz continuity of \( \nabla_{\theta^Q} L_{\mathrm{reg}} \), we get
\[
\left\| \nabla_{\theta^Q} L_{\mathrm{reg}}(\theta_k^Q) - \nabla_{\theta^Q} L_{\mathrm{reg}}(\theta^{Q^*}) \right\| \leq L \left\| e_k \right\|,
\]
where \( L \) is the Lipschitz constant. We can then bound the cross term
\[
\mathbb{E}\left[ e_k^T \nabla_{\theta^Q} L_{\mathrm{reg}}(\theta_k^Q) \right] \geq m \mathbb{E}\left[ \left\| e_k \right\|^2 \right].
\]
Moreover, the norm of the gradient is bounded, \( \left\| \nabla_{\theta^Q} L_{\mathrm{reg}}(\theta_k^Q) \right\| \leq G \), giving:
\[
\mathbb{E}\left[ \left\| \nabla_{\theta^Q} L_{\mathrm{reg}}(\theta_k^Q) \right\|^2 \right] \leq G^2.
\]
Combining these bounds, we have
\[
\mathbb{E}\left[\left\| e_{k+1} \right\|^2\right] \leq \mathbb{E}\left[\left\| e_k \right\|^2\right] - 2\alpha_k m \mathbb{E}\left[\left\| e_k \right\|^2\right] + \alpha_k^2 G^2.
\]
Simplifying, we get
$
\mathbb{E}\left[\left\| e_{k+1} \right\|^2\right] \leq \left(1 - 2\alpha_k m\right) \mathbb{E}\left[\left\| e_k \right\|^2\right] + \alpha_k^2 G^2.
$
Define \( \phi_k = \mathbb{E}\left[\left\| e_k \right\|^2\right] \), giving
\[
\phi_{k+1} \leq \left(1 - \frac{2am}{k}\right) \phi_k + \frac{a^2 G^2}{k^2}.
\]
By applying a standard result for such recursive sequences, we conclude that
$
\phi_k \leq \frac{C}{k}.
$
for some constant \( C \) dependent on the initial error and the parameters \( a, m, G \). Thus, 
$
\mathbb{E}\left[\left\| e_k \right\|^2\right] \leq \frac{C}{k},
$
implying the mean squared error decreases at a rate of \( \mathcal{O}\left(\frac{1}{k}\right) \). Taking square roots, we have:
\[
\mathbb{E}\left[\left\| e_k \right\|\right] \leq \frac{\sqrt{C}}{\sqrt{k}} = \mathcal{O}\left(\frac{1}{\sqrt{k}}\right),
\]
showing that the critic's parameter estimates converge to the optimal parameters at a rate of \( \mathcal{O}\left(\frac{1}{\sqrt{k}}\right) \).
\qed
\end{proof}

\begin{theorem}[Regularization-Induced Stability]
 Incorporating the Wasserstein regularization term \( \lambda_k W_{\varepsilon}\left(\hat{Q}_k, \hat{Q}_{k-1}\right) \) into the critic loss function enhances the stability of the critic's updates by penalizing large deviations between successive \( Q \)-value estimates.   
\end{theorem}
\begin{proof}
To show that the inclusion of the Wasserstein regularization term enhances the stability of the critic’s updates, we analyze how this term influences the critic’s loss function and the parameter updates. Specifically, we aim to demonstrate that the regularization term acts as a penalty for large deviations between consecutive \( Q \)-value estimates, thereby reducing the variance of updates and promoting smoother learning. We begin by defining the critic network and the loss functions used. Let \( Q(s,a; \theta^Q_k) \) represent the critic network at iteration \( k \) with parameters \( \theta^Q_k \). The standard temporal difference (TD) error loss is given by
$
L_{\text{TD}}(\theta^Q_k) = \mathbb{E}_{(s,a,r,s') \sim \mathcal{D}} \left[ \left( Q(s,a; \theta^Q_k) - y \right)^2 \right],
$
where \( y = r + \gamma Q(s', \mu(s'; \theta^{\mu^-}); \theta^{Q^-}) \) represents the target \( Q \)-value. Next, we introduce the Wasserstein regularization term. Consider \( \hat{Q}_k \) and \( \hat{Q}_{k-1} \) as the empirical distributions of \( Q \)-value estimates at iterations \( k \) and \( k - 1 \), respectively. The Sinkhorn distance \( W_{\varepsilon}(\hat{Q}_k, \hat{Q}_{k-1}) \) is used to approximate the Wasserstein distance between these distributions. The regularized critic loss function is defined as
\[
L_{\text{reg}}(\theta^Q_k) = L_{\text{TD}}(\theta^Q_k) + \lambda_k W_{\varepsilon}(\hat{Q}_k, \hat{Q}_{k-1}),
\]
where \( \lambda_k \geq 0 \) is an adaptive regularization parameter. The Sinkhorn distance \( W_{\varepsilon}(\hat{Q}_k, \hat{Q}_{k-1}) \) measures the discrepancy between \( \hat{Q}_k \) and \( \hat{Q}_{k-1} \). By incorporating \( \lambda_k W_{\varepsilon}(\hat{Q}_k, \hat{Q}_{k-1}) \) into the loss function, large shifts in the \( Q \)-value distribution are penalized, enforcing temporal consistency in the learning process. The gradient of the regularized loss function with respect to \( \theta^Q_k \) is given by
\[
\nabla_{\theta^Q_k} L_{\text{reg}}(\theta^Q_k) = \nabla_{\theta^Q_k} L_{\text{TD}}(\theta^Q_k) + \lambda_k \nabla_{\theta^Q_k} W_{\varepsilon}(\hat{Q}_k, \hat{Q}_{k-1}).
\]
Here, the regularization gradient \( \lambda_k \nabla_{\theta^Q_k} W_{\varepsilon}(\hat{Q}_k, \hat{Q}_{k-1}) \) acts to counterbalance abrupt changes in \( Q \)-values. Considering the variance of parameter updates, the update rule for \( \theta^Q_k \) with learning rate \( \eta_k \) is given by
$
\theta^Q_{k+1} = \theta^Q_k - \eta_k \nabla_{\theta^Q_k} L_{\text{reg}}(\theta^Q_k).
$
The variance of these updates depends on the variance of the gradients:
\[
\text{Var}\left( \Delta \theta^Q_k \right) = \eta_k^2 \text{Var}\left( \nabla_{\theta^Q_k} L_{\text{reg}}(\theta^Q_k) \right).
\]
Let \( g_{\text{TD}} = \nabla_{\theta^Q_k} L_{\text{TD}}(\theta^Q_k) \) and \( g_{\text{W}} = \nabla_{\theta^Q_k} W_{\varepsilon}(\hat{Q}_k, \hat{Q}_{k-1}) \). The total gradient variance is expressed as
\[
\text{Var}(g_{\text{TD}} + \lambda_k g_{\text{W}}) = \text{Var}(g_{\text{TD}}) + \lambda_k^2 \text{Var}(g_{\text{W}}) + 2 \lambda_k \text{Cov}(g_{\text{TD}}, g_{\text{W}}).
\]
Assuming that the covariance \( \text{Cov}(g_{\text{TD}}, g_{\text{W}}) \leq 0 \) (because \( g_{\text{W}} \) generally opposes changes suggested by \( g_{\text{TD}} \)), the overall variance is reduced. Although the regularization introduces a bias by discouraging large shifts in \( Q \)-values, it simultaneously reduces the gradient variance. An adaptive strategy for selecting \( \lambda_k \) helps ensure that this bias diminishes over time as performance stabilizes, balancing exploration and exploitation. The argument for stability rests on specific assumptions. First, assume that the critic network \( Q(s,a; \theta^Q) \) is Lipschitz continuous with respect to \( \theta^Q \). Additionally, assume that the gradients \( \nabla_{\theta^Q} Q(s,a; \theta^Q) \) and \( \nabla_{\theta^Q} W_{\varepsilon}(\hat{Q}_k, \hat{Q}_{k-1}) \) are bounded, with constants \( G_{\text{TD}} \) and \( G_{\text{W}} \) such that
\[
\| g_{\text{TD}} \| \leq G_{\text{TD}}, \quad \| g_{\text{W}} \| \leq G_{\text{W}}.
\]
The inclusion of the regularization term ensures that
\[
\| \nabla_{\theta^Q_k} L_{\text{reg}}(\theta^Q_k) \| \leq G_{\text{TD}} + \lambda_k G_{\text{W}},
\]
thus controlling the magnitude of updates. By limiting step sizes, we achieve smoother updates, which lead to more consistent \( Q \)-value estimates over iterations. Lastly, the Sinkhorn distance \( W_{\varepsilon} \) has useful properties. It is differentiable and provides gradients that help align the distributions \( \hat{Q}_k \) and \( \hat{Q}_{k-1} \), further reducing discrepancies between successive \( Q \)-value distributions. This alignment contributes to the empirical stability of the critic's updates.
\qed
\end{proof}

\begin{theorem}[Accelerated Convergence via Enhanced Contraction of the Regularized Bellman Operator]
%\begin{theorem}[Accelerated Convergence by Bellman Operator Contraction]
Incorporating adaptive Wasserstein regularization into the critic's loss function enhances the contraction property of the Bellman operator in the Wasserstein metric space. Specifically, under standard assumptions, the regularized Bellman operator \( \mathcal{T}_\lambda \) becomes a contraction with a contraction factor \( \gamma_\lambda = \gamma (1 - c \lambda) \), where \( \gamma \in (0,1) \) is the discount factor, \( \lambda > 0 \) is the regularization parameter, and \( c > 0 \) is a constant related to the effect of the regularization.  
\end{theorem}
\begin{proof}
    We aim to show that the adaptive Wasserstein regularization modifies the Bellman operator to become a stronger contraction mapping in the Wasserstein metric space, thereby accelerating the convergence of the value function estimates. Consider an MDP defined by the tuple \( (\mathcal{S}, \mathcal{A}, P, r, \gamma) \), where \( \mathcal{S} \) is the state space, \( \mathcal{A} \) is the action space, \( P \) denotes the transition probabilities, and \( r: \mathcal{S} \times \mathcal{A} \rightarrow \mathbb{R} \) is the reward function. The objective is to estimate the value function \( Q^\pi(s, a) \) for a given policy \( \pi \), using an actor-critic framework with adaptive Wasserstein regularization. Let \( \mathcal{T} \) denote the standard Bellman operator, defined as
\[
\mathcal{T} Q(s, a) = r(s, a) + \gamma \mathbb{E}_{s' \sim P, a' \sim \pi} \left[ Q(s', a') \right].
\]
It is well-known that \( \mathcal{T} \) is a contraction mapping with contraction factor \( \gamma \) in the supremum norm, meaning
\[
\| \mathcal{T} Q_1 - \mathcal{T} Q_2 \|_\infty \leq \gamma \| Q_1 - Q_2 \|_\infty.
\]
We extend this contraction property to the Wasserstein metric \( W_p \) for \( p \geq 1 \) under certain conditions on the transition kernel and the policy. Now, consider the regularized Bellman operator \( \mathcal{T}_\lambda \), which incorporates adaptive Wasserstein regularization
\[
\mathcal{T}_\lambda Q(s, a) = \mathcal{T} Q(s, a) - \lambda \frac{\partial W_\varepsilon(\hat{Q}_k, \hat{Q}_{k-1})}{\partial Q(s, a)},
\]
where \( \lambda > 0 \) is the regularization parameter, and \( W_\varepsilon \) is the entropic regularization of the Wasserstein distance between successive value function distributions. The objective is to show that \( \mathcal{T}_\lambda \) has an enhanced contraction property in the Wasserstein metric. To analyze the effect of the regularization, consider the difference between the applications of the regularized Bellman operator to two value functions \( Q_1 \) and \( Q_2 \)
\begin{align*}
\mathcal{T}_\lambda Q_1(s, a) - \mathcal{T}_\lambda Q_2(s, a) 
&= \mathcal{T} Q_1(s, a) - \mathcal{T} Q_2(s, a) \\
&\quad - \lambda \left( 
\frac{\partial W_\varepsilon(\hat{Q}_{1,k}, \hat{Q}_{1,k-1})}{\partial Q_1(s, a)} 
- \frac{\partial W_\varepsilon(\hat{Q}_{2,k}, \hat{Q}_{2,k-1})}{\partial Q_2(s, a)} 
\right).
\end{align*}

% \[
% \mathcal{T}_\lambda Q_1(s, a) - \mathcal{T}_\lambda Q_2(s, a) = \mathcal{T} Q_1(s, a) - \mathcal{T} Q_2(s, a) - \lambda \left( \frac{\partial W_\varepsilon(\hat{Q}_{1,k}, \hat{Q}_{1,k-1})}{\partial Q_1(s, a)} - \frac{\partial W_\varepsilon(\hat{Q}_{2,k}, \hat{Q}_{2,k-1})}{\partial Q_2(s, a)} \right).
% \]
The first term, \( \mathcal{T} Q_1 - \mathcal{T} Q_2 \), satisfies the standard contraction property
\[
| \mathcal{T} Q_1(s, a) - \mathcal{T} Q_2(s, a) | \leq \gamma \mathbb{E}_{s', a'} \left[ | Q_1(s', a') - Q_2(s', a') | \right].
\]
The second term involves the derivative of the Wasserstein distance and reflects the impact of regularization. The regularization term penalizes large discrepancies between \( Q_1 \) and \( Q_2 \) across iterations, thereby inducing a smoothing effect. By properties of the Sinkhorn distance and Lipschitz continuity, we have
\[
\left| \frac{\partial W_\varepsilon(\hat{Q}_{1,k}, \hat{Q}_{1,k-1})}{\partial Q_1(s, a)} - \frac{\partial W_\varepsilon(\hat{Q}_{2,k}, \hat{Q}_{2,k-1})}{\partial Q_2(s, a)} \right| \geq c_1 | Q_1(s, a) - Q_2(s, a) |,
\]
for some constant \( c_1 > 0 \). Combining these, we obtain
\[
| \mathcal{T}_\lambda Q_1(s, a) - \mathcal{T}_\lambda Q_2(s, a) | \leq \gamma \mathbb{E}_{s', a'} \left[ | Q_1(s', a') - Q_2(s', a') | \right] - \lambda c_1 | Q_1(s, a) - Q_2(s, a) |.
\]
Taking the expectation over the next states and actions and applying the Lipschitz continuity of the transition kernel, we get
\[
| \mathcal{T}_\lambda Q_1(s, a) - \mathcal{T}_\lambda Q_2(s, a) | \leq \left( \gamma - \lambda c_1 \right) | Q_1(s, a) - Q_2(s, a) |.
\]
Defining \( c = \frac{c_1}{\gamma} \), we have
\[
| \mathcal{T}_\lambda Q_1(s, a) - \mathcal{T}_\lambda Q_2(s, a) | \leq \gamma (1 - c \lambda) | Q_1(s, a) - Q_2(s, a) |,
\]
which implies that the regularized Bellman operator \( \mathcal{T}_\lambda \) is a contraction with contraction factor \( \gamma_\lambda = \gamma (1 - c \lambda) \). Since \( \gamma_\lambda < \gamma \) for \( \lambda > 0 \) and \( c > 0 \), the contraction factor is reduced, leading to accelerated convergence. The convergence rate of the value function estimates \( \{ Q_k \} \) to the fixed point \( Q^* \) is given by
\[
W_p(Q_{k+1}, Q^*) \leq \gamma_\lambda W_p(Q_k, Q^*),
\]
where \( W_p \) denotes the Wasserstein distance. Consequently, the sequence \( \{ Q_k \} \) converges faster to \( Q^* \), enhancing the overall learning efficiency of the actor-critic algorithm.
\qed
\end{proof}
\begin{algorithm}[H]
\SetAlgoLined
\SetKwInOut{Input}{Input}\SetKwInOut{Output}{Output}
\SetKwComment{Comment}{$\triangleright$\ }{}

\caption{Wasserstein Adaptive Value Estimation RL}

\Input{Initial parameters $\theta^\pi$, $\theta^Q$, $\lambda_{\max}$}
\Output{Updated actor $\pi(a \mid s; \theta^\pi)$ and critic $Q^\pi(s, a; \theta^Q)$ parameters}

\BlankLine
\textbf{Initialization:} \\
- Initialize actor $\pi(a \mid s; \theta^\pi)$ and critic $Q^\pi(s, a; \theta^Q)$\;
- Set regularization parameter $\lambda_1 \gets \lambda_{\max}$\;

\BlankLine
\For{each episode $k$}{
    \Comment{Collect data}
    Collect trajectories $\left\{ \left( s_t, a_t, r_t, s_{t+1} \right) \right\}$ by interacting with the environment\;
    
    \BlankLine
    \Comment{Update the critic}
    Update critic parameters $\theta^Q$ by minimizing the regularized loss $L_{\mathrm{reg}}(\theta^Q)$\;
    
    \BlankLine
    \Comment{Update the actor}
    Update actor parameters $\theta^\pi$ using policy gradient\;
    
    \BlankLine
    \Comment{Compute and adjust}
    Compute cumulative reward $R_k$ and update the running average $\bar{R}_k$\;
    Adjust the regularization parameter $\lambda_k$ based on $\bar{R}_k$\;
}
\label{alg:wave-rl}
\end{algorithm}

\section{Numerical Results}
\begin{figure}[b]
    \centering
    \begin{tabular}{@{}c@{\hspace{2mm}}c@{\hspace{2mm}}c@{}}
        \includegraphics[width=0.32\textwidth]{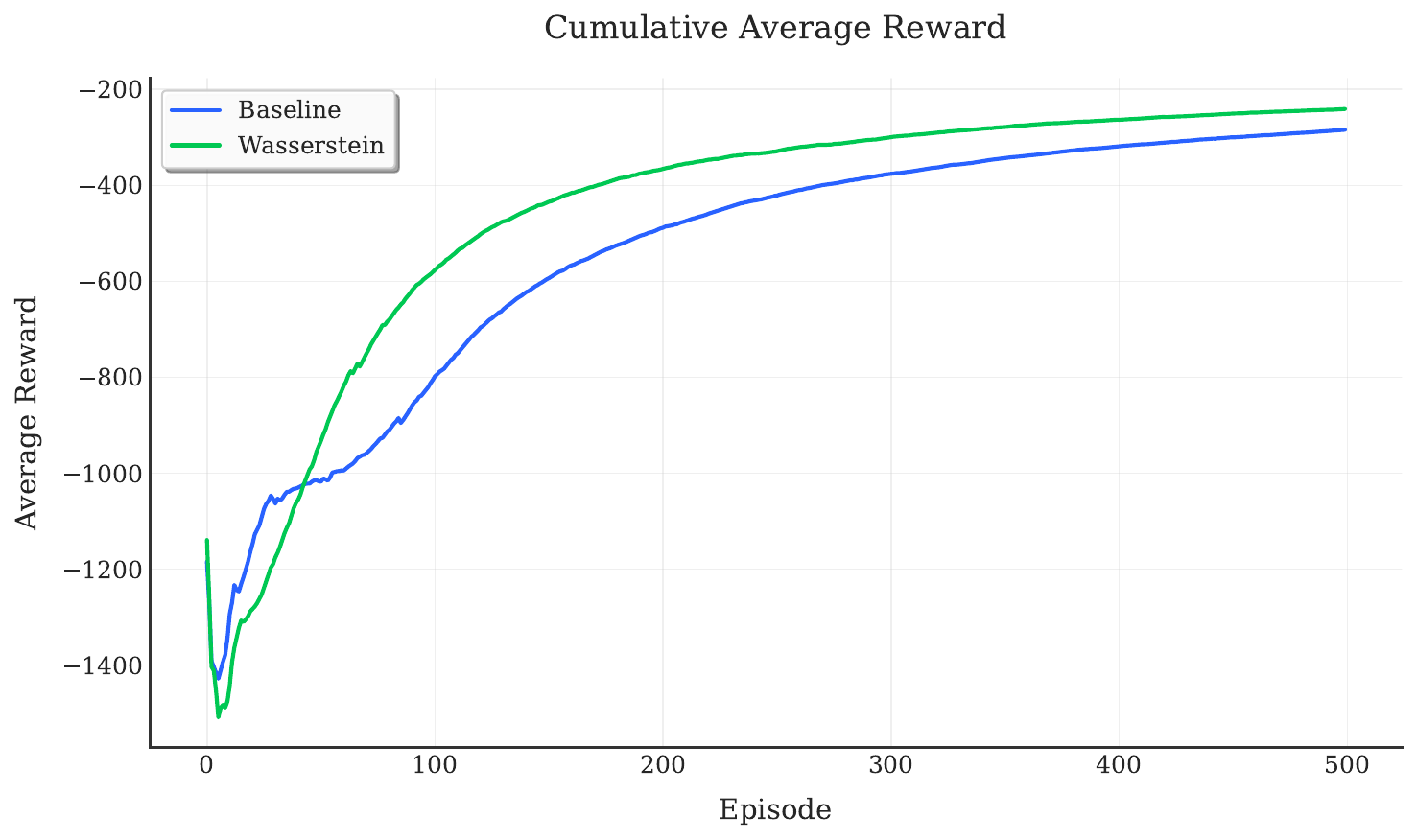} &
        \includegraphics[width=0.32\textwidth]{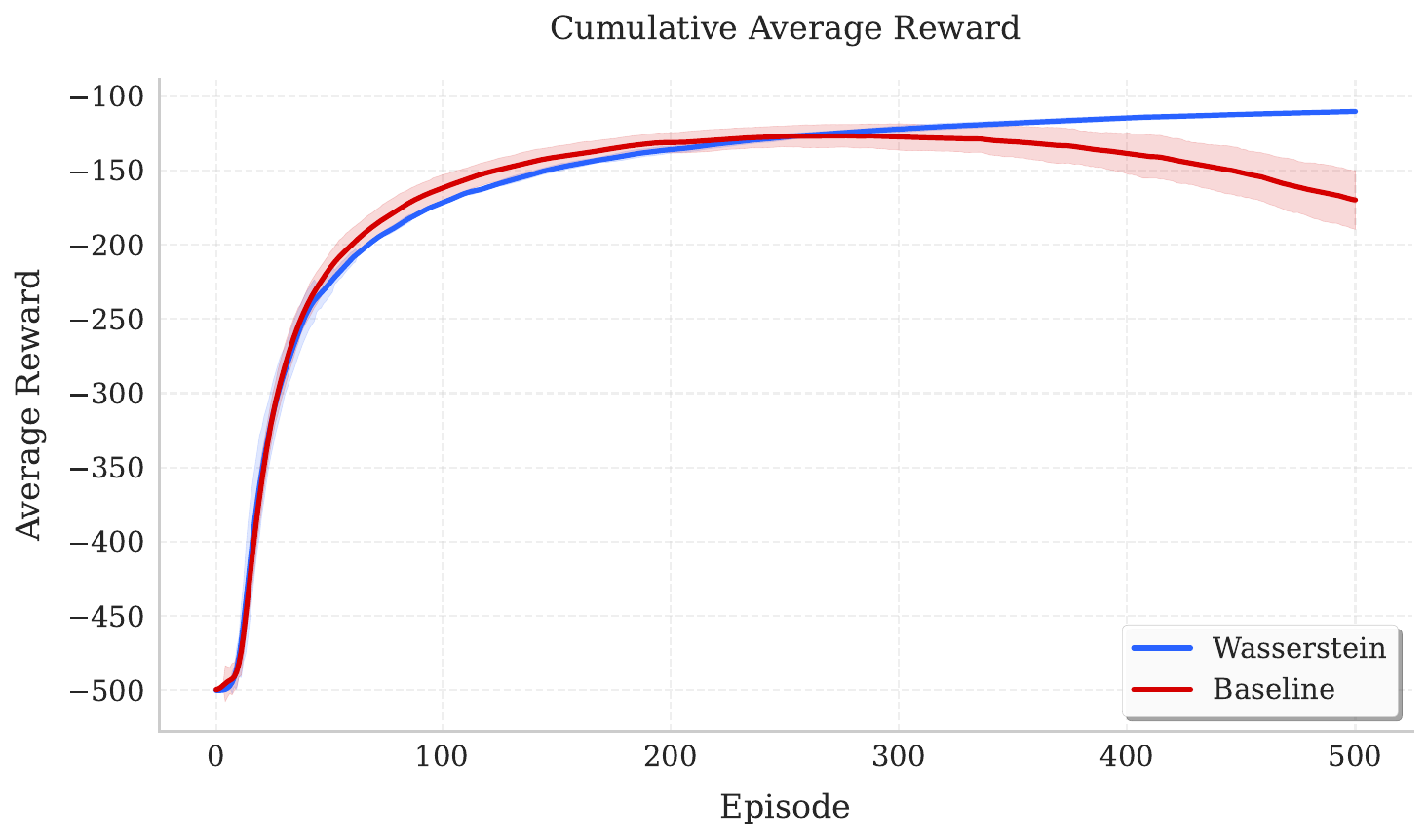} &
        \includegraphics[width=0.32\textwidth]{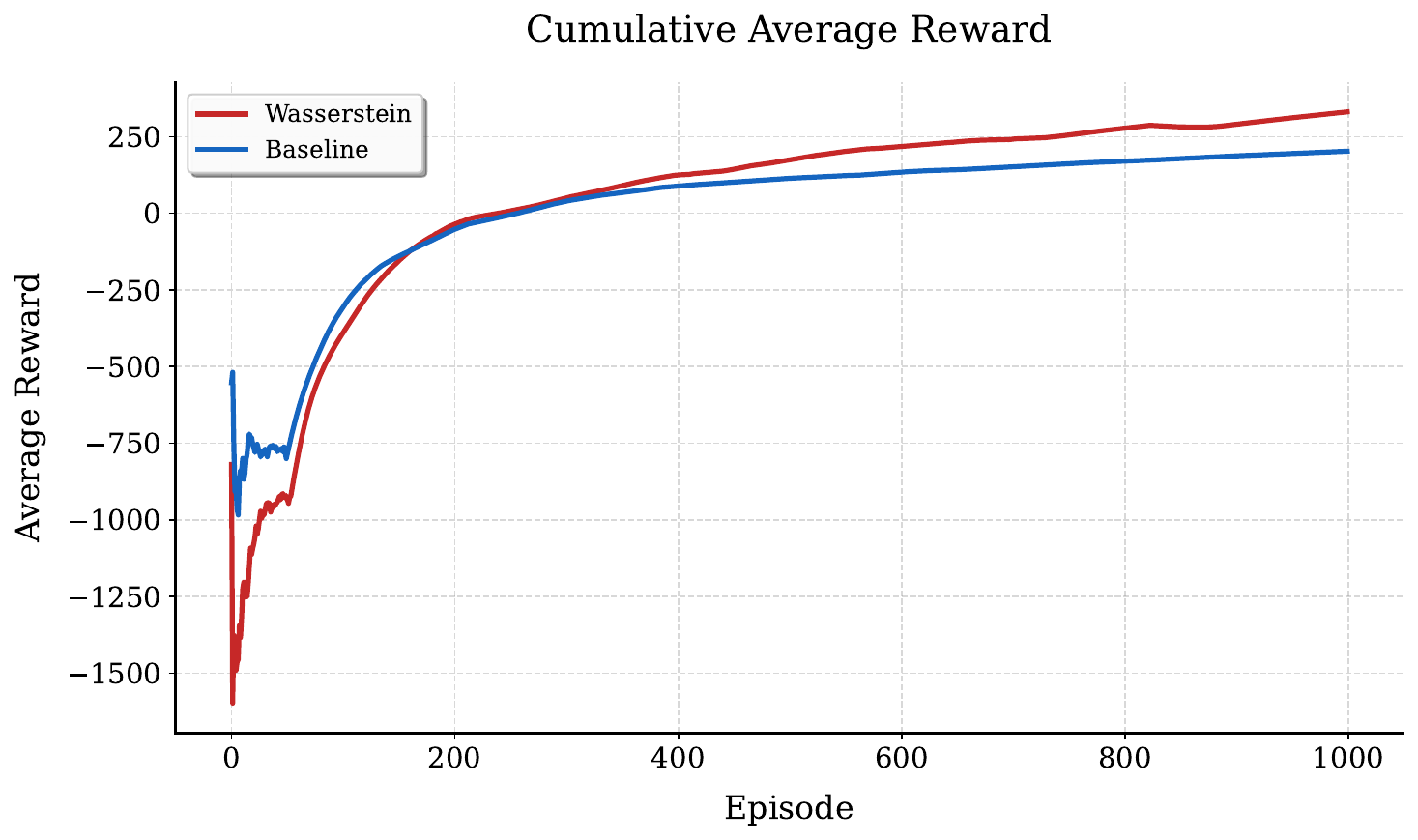} \\
    \end{tabular}
    \caption{Comparison of cumulative average rewards between WAVE and baseline across three continuous control environments: Inverted Pendulum (left), Acrobot (middle), and 2D Robot Navigation (right).}
    \label{fig:three_figures}
\end{figure}
We evaluated WAVE through experiments in three continuous control environments: \textit{Inverted Pendulum}, \textit{Acrobot}, and a custom \textit{2D Robot Navigation} task, compared to TD3 (twin delayed deep deterministic policy gradient \cite{fujimoto2018addressing}) as the baseline actor-critic algorithm. All experiments employ identical network architectures, utilizing multi-layer perceptrons with three hidden layers (256 units each) and LayerNorm for both actor and critic networks. The critic implements a twin Q-network architecture to mitigate overestimation bias. We maintain consistent hyperparameters across all environments, with learning rates of \(3 \times 10^{-4}\), batch size of 256, and discount factor of 0.99. The algorithm parameters include initial regularization \(\lambda_{\text{max}} = 2.0\), minimum regularization \(\lambda_{\text{min}} = 0.3\), with the Sinkhorn approximation parameter \(\epsilon = 0.005\). Fig. \ref{fig:three_figures} shows the comparison of cumulative average rewards between WAVE and baseline across three case studies. The results on the Inverted Pendulum environment demonstrate WAVE's superior convergence properties. 

\begin{wrapfigure}{r}{0.6\textwidth} % 'r' for right, width of 0.4\textwidth
    \centering
    \includegraphics[width=0.6\textwidth]{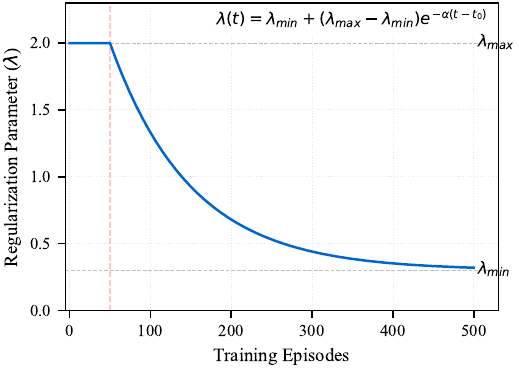} % Adjust width as needed
    \caption{\scriptsize Evolution of the adaptive regularization parameter during training on the Inverted Pendulum environment.}
    \label{fig:reg}
\end{wrapfigure}
The Acrobot environment results further validate WAVE's effectiveness. The proposed method achieves a final average reward of approximately -100, outperforming the baseline's reward. In the custom 2D Robot Navigation environment WAVE exhibits the enhanced final performance. The empirical results show an approximately 25\% improvement in cumulative average reward compared to baseline. Analysis of the regularization parameter \(\lambda_k\) throughout training in Fig. \ref{fig:reg} shows adaptation to learning progress. During initial exploration, the high regularization value stabilizes learning, while gradual reduction allows for refined policy adjustments as performance improves. 

\section{Discussion}
Although the proposed algorithm offers improvements in convergence for actor-critic RL, it does have certain limitations that should be acknowledged. One concern is the computational overhead associated with computing the Sinkhorn approximation of the Wasserstein distance, especially in environments with high-dimensional state and action spaces. Additionally, the adaptive regularization parameter $\lambda_k$ relies on the cumulative reward to adjust its value, which assumes that the reward signal is a reliable indicator of the agent's performance. In environments with sparse or delayed rewards, this assumption may not hold. Furthermore, the theoretical convergence guarantees provided are contingent upon specific assumptions, such as the Lipschitz continuity of the critic network and the strong convexity of the regularized loss function, which may not be universally applicable across all neural network architectures or RL tasks. Finally, the method introduces additional hyperparameters, such as $\lambda_{\max }, \lambda_{\min }$, and the decay rate $\alpha$, which require tuning to achieve optimal performance. 

\section{Conclusion}
This study proposed WAVE, a framework designed to enhance stability in actor-critic RL through adaptive Wasserstein regularization. The proposed method improves convergence and stability by penalizing large deviations in value estimates while dynamically adjusting regularization parameters based on agent performance. Theoretical analysis guarantees an $\mathcal{O}\left(\frac{1}{k}\right)$ convergence rate for the critic's mean squared error, whereas empirical evaluations across continuous control environments validate the approach's superior overall performance. WAVE demonstrates a compelling integration of optimal transport theory into RL, offering a solution to instability and value approximation challenges in complex tasks.

\subsubsection{\discintname}
The authors have no competing interests to declare that they are
relevant to the content of this article. 

\bibliographystyle{splncs04}
\bibliography{ref}

\end{document}